\documentclass[10pt,twocolumn,letterpaper]{article}

\usepackage[pagenumbers]{cvpr}

\usepackage{url}
\usepackage{booktabs}
\usepackage{multirow}
\usepackage{graphicx}
\usepackage{placeins}

\definecolor{cvprblue}{rgb}{0.21,0.49,0.74}
\usepackage[pagebackref,breaklinks,colorlinks,allcolors=cvprblue]{hyperref}

\def\confName{CVPR}
\def\confYear{2027}

\title{Motion-Aware Reward Modeling for Instruction-Guided Video Editing}

\author{
Li Li$^1$, Xiangbo Gao$^2$, Bo Ni$^3$, Samyadeep Basu$^4$, Viet Dac Lai$^4$, Jing Shi$^4$,\\
Seunghyun Yoon$^4$, Gang Wu$^4$, Yu Wang$^5$, Thien Huu Nguyen$^6$, Tyler Derr$^3$,\\
Franck Dernoncourt$^4$, Zhengzhong Tu$^2$, Yue Zhao$^1$\\
$^1$University of Southern California \quad $^2$Texas A\&M University\\
$^3$Vanderbilt University \quad $^4$Adobe\\
$^5$University of Georgia \quad $^6$Amazon
}

\begin{document}
\maketitle

\begin{abstract}
Reward models for instruction-guided video editing can favor temporal consistency at the expense of motion. We study this failure, termed \emph{motion collapse}, using controlled motion-suppression variants of real edits. On VE-Bench, the evaluated composite reward prefers a motion-suppressed candidate in $97\%$ of cases. Learned video rewards and a VLM judge also struggle with mild suppression, while latent optimization and reward-guided fine-tuning exploit this weakness to increase reward as motion declines. A linear probe of our base reward backbone performs near chance on mild suppression, motivating an explicit inter-frame dynamics input. We propose a motion-aware reward model that combines an RGB frame-difference stream with preference training against motion-suppressed and fabricated-motion negatives. A distillation anchor preserves alignment with human quality judgments. On the VE-Bench motion ladder, our model reduces the rate of selecting motion-suppressed candidates to $0\%$ while achieving a Spearman correlation of $0.57$ with human quality ratings. It also detects unseen suppression variants and preserves motion during reward-guided fine-tuning on two editing pipelines. These results support explicitly modeling motion in rewards for video editing.
\end{abstract}

\section{Introduction}
\label{sec:intro}

Instruction-guided video editing must balance requested changes with preservation of the source content. When an instruction changes appearance but leaves the action unspecified, changing a person's clothing should not make them stop walking. Rewards used to optimize editing models~\citep{viva2026} therefore need to assess both edit quality and motion preservation. Yet temporal consistency alone is insufficient: adjacent frames can become more similar because motion has been removed. Prior work has identified low-motion biases in video evaluation and preference optimization~\citep{devil2024,densedpo2025,longcat2025}. We investigate how this problem manifests in video-editing rewards and how to make these rewards sensitive to motion degradation.

We construct a controlled \emph{motion ladder} from each generated edit by repeating frames over progressively longer intervals, ending with a fully frozen clip. On edits from VE-Bench~\citep{vebench2025}, the composite reward we evaluate ranks a motion-suppressed candidate above the original edit in $97\%$ of cases; removing its temporal-consistency component substantially reduces this preference. The tested learned video rewards and a VLM judge also struggle with mild suppression. Human raters, however, overwhelmingly prefer the original edits, confirming that these reward rankings miss a perceptually important distinction.

This mismatch also affects optimization. In our experiments, both latent optimization and reward-guided fine-tuning can increase reward while reducing motion, even as measured prompt adherence improves. We refer to this failure in video editing as \emph{motion collapse}. It motivates testing rewards under optimization in addition to evaluating their agreement with human ratings on unmodified outputs.

\begin{figure*}[t]
\centering
\includegraphics[width=\linewidth]{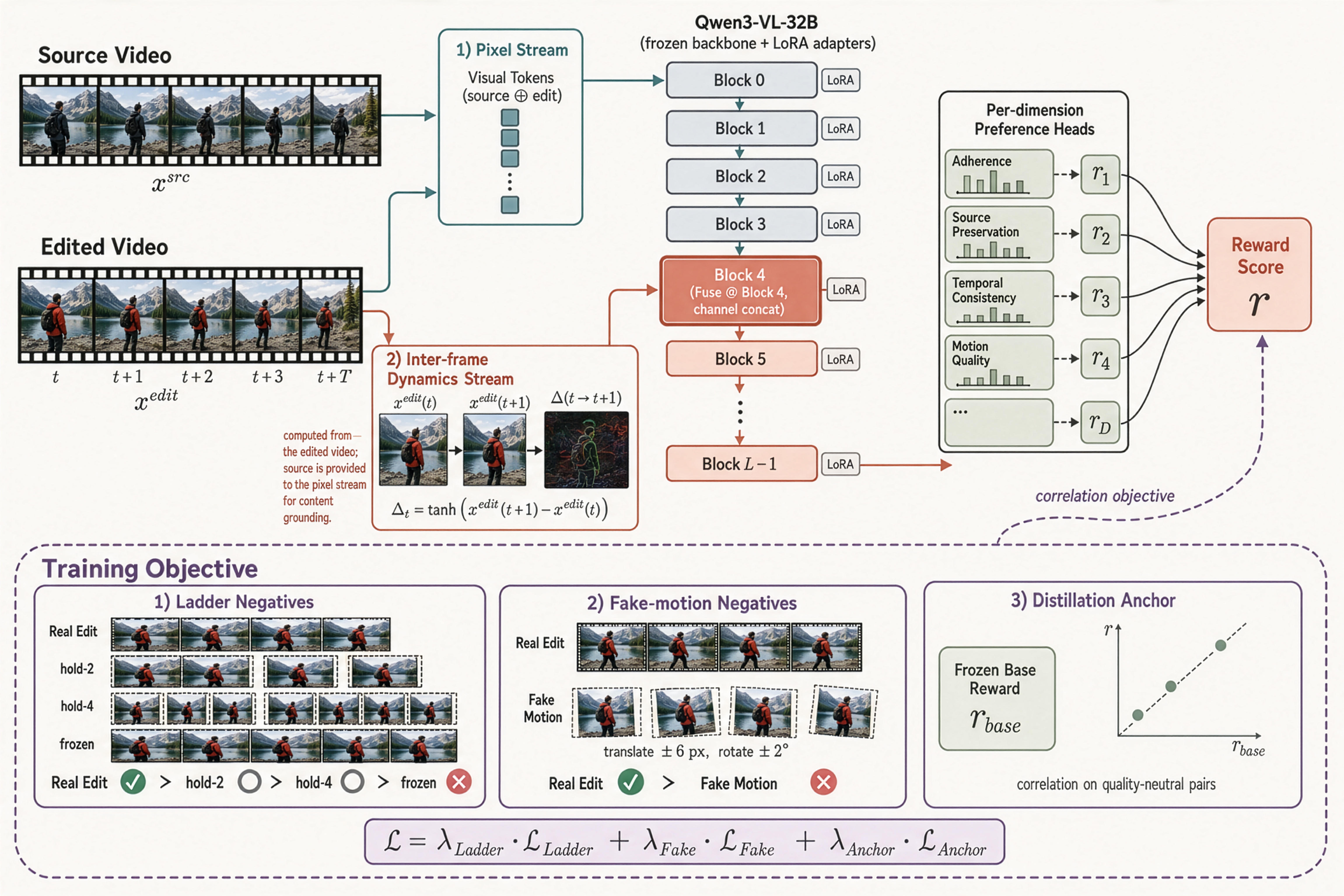}
\caption{Motion-aware reward model. The pixel stream encodes source and edited videos, while a dynamics stream computes tanh-normalized RGB differences of the edited video. Their features are fused at block 4 of the Qwen3-VL-32B backbone and adapted with LoRA. Training combines pairwise preferences over original edits and motion-suppressed or fabricated-motion negatives with a distillation anchor to the base preference reward.}
\label{fig:architecture}
\end{figure*}

To guide reward design, we probe the representations of our base preference model. A linear classifier distinguishes original edits from fully frozen clips but performs near chance on mildly suppressed variants. Together with the limited gains from fine-tuning without motion input, this finding motivates an explicit dynamics stream. Our motion-aware reward model combines RGB frame differences with source and edited-video features (Figure~\ref{fig:architecture}). Preference training contrasts original edits with both motion-suppressed and fabricated-motion negatives, teaching the model to reject motion degradation without simply rewarding more movement. A distillation anchor retains the base model's quality preferences, keeping motion sensitivity compatible with broader editing-quality assessment.

On the VE-Bench motion ladder, our reward selects the original edit in every evaluated case and achieves a Spearman correlation of $0.57$ with human quality ratings. It also detects suppression patterns absent from training. When used for reward-guided fine-tuning on two editing pipelines, it maintains motion close to its initial level. These results support evaluating motion sensitivity and quality alignment jointly when designing rewards for video editing.

Our contributions are threefold:
\begin{itemize}
    \item We diagnose motion collapse in video-editing rewards through controlled temporal perturbations, human preference comparisons, and reward-optimization experiments.
    \item We develop a motion-aware reward model that combines explicit inter-frame dynamics, motion-suppressed and fabricated-motion negatives, and preference-preserving distillation.
    \item We evaluate its robustness to unseen motion-suppression patterns, alignment with human quality judgments, and ability to preserve motion during reward-guided fine-tuning.
\end{itemize}

\section{Related Work}
\label{sec:related}

\noindent \textbf{Visual reward models.} EditReward and EditScore assess instruction-guided image edits~\citep{editreward2026,editscore2026}, without explicitly modeling inter-frame dynamics. VideoScore and VideoReward learn multi-dimensional video quality scores~\citep{videoscore2024,videoreward2025}; VisionReward uses binary quality judgments~\citep{visionreward2024}, while VideoScore2 adds reasoning-based evaluation~\citep{videoscore22025}. We examine whether preference agreement translates into sensitivity to motion suppression.

\noindent \textbf{Video evaluation.} VE-Bench provides human-rated edits and a learned quality assessor~\citep{vebench2025}; IVEBench includes source-relative trajectory fidelity~\citep{ivebench2026}. VBench separates motion smoothness from dynamic degree~\citep{vbench2024}, and DEVIL identifies inflated quality scores for low-dynamic videos~\citep{devil2024}. Building on these distinctions, we test controlled motion suppression and its optimization consequences. Our composite is distinct from the complete VE-Bench and IVEBench evaluators. Related evaluation challenges arise in scientific chart editing, where pixel similarity can miss semantic errors~\citep{li2025chartsimageschallengesscientific}.

\noindent \textbf{Reward optimization and robustness.} VIVA's Edit-GRPO combines CLIP and PickScore rewards for video editing~\citep{viva2026}. DenseDPO mitigates low-motion preference bias through motion-aligned pairs and segment-level supervision~\citep{densedpo2025}. Reward overoptimization can degrade the intended objective~\citep{overoptimization2023}: LongCat-Video reports reward-induced static outputs and counteracts them with multiple rewards~\citep{longcat2025}, while SoliReward improves robustness through annotation, architecture, and ranking-loss design~\citep{solireward2026}. We investigate motion sensitivity within the reward model, testing optimization behavior through latent updates and fine-tuning adapted from DRaFT~\citep{draft2024}.

\noindent \textbf{Broader reliability and evaluation.} Representation bias is studied through predicate-annotation correction~\citep{limm,Li_Ji_Wu_Li_Qin_Wei_Zimmermann_2024,10447193} and counterfactual analysis of VLM hallucinations~\citep{ye2025towards,li-etal-2025-treble}. Complementary work addresses distribution shifts through multimodal prototypes~\citep{Li_2025_CVPR}, on-device video OOD detection~\citep{li2025secureondevicevideoood}, and diffusion-based cross-domain detection~\citep{li2026geometrydensityfewshotcrossdomain}. In language models, defense training can learn surface heuristics~\citep{li2026defensespromptattackslearn} and impair agent competence~\citep{li2026autonomytaxdefensetraining}, illustrating the need to assess capabilities beyond the training objective. Task-specific benchmarks also examine personalized conversational context~\citep{li2025personalizedconversationalbenchmarksimulating} and agent privilege boundaries~\citep{li2026fortisbenchmarkingoverprivilegeagent}. These studies provide broader context for our evaluation of motion preservation under reward optimization.

\noindent \textbf{Motion representations.} Two-stream networks separate RGB appearance and optical-flow inputs~\citep{twostream2014}; Temporal Difference Networks encode motion through frame and feature differences~\citep{tdn2021}. We incorporate RGB differences into a reward backbone and train with motion-suppressed and fabricated-motion negatives. RAFT~\citep{raft2020} supports motion measurement and our external baseline; the learned dynamics stream requires no flow estimator.

\begin{table*}[t]
\centering
\small
\caption{Mean scores across the VE-Bench motion ladder and top-1 capture rates. The composite combines appearance and temporal terms; the appearance-only variant removes the latter. VideoReward is reported overall and by head.}
\label{tab:diagnosis}
\begin{tabular}{lccccc}
\toprule
Reward & genuine & hold-2 & hold-4 & frozen & capture $\downarrow$ \\
\midrule
\multicolumn{6}{l}{\textit{Field metrics}} \\
RAFT warping error ($\downarrow$) & 33.20 & 23.39 & 16.08 & 0.03 & 100\% \\
DINO temporal consistency ($\uparrow$) & 0.869 & 0.919 & 0.950 & 1.000 & 100\% \\
\textbf{Field composite} (appearance $+$ temporal) & \multicolumn{4}{c}{\multirow{2}{*}{$\bigg\{$~\textit{z-normalized weighted combination}~$\bigg\}$}} & \textbf{97\%} \\
Field composite (appearance only) & \multicolumn{4}{c}{} & 62\% \\
\midrule
\multicolumn{6}{l}{\textit{Learned video reward models}} \\
VideoScore-v1.1 \citep{videoscore2024} & 3.378 & 3.357 & 3.357 & 3.392 & 80\% \\
VideoReward \citep{videoreward2025} & $-1.54$ & $-1.38$ & $-1.43$ & $-1.51$ & 76\% \\
\quad \textbf{motion-quality head} & $\mathbf{-0.65}$ & $\mathbf{-0.33}$ & $\mathbf{-0.33}$ & $\mathbf{-0.13}$ & \textbf{98\%} \\
\quad text-alignment head (control) & $-0.17$ & $-0.49$ & $-0.51$ & $-0.76$ & 22\% \\
\bottomrule
\end{tabular}
\end{table*}

\section{Motion Collapse}
\label{sec:motion-collapse}

We test reward sensitivity to temporal degradation with a \emph{motion ladder}: the original generated edit, hold-2 and hold-4 variants that repeat every second or fourth frame, and a frozen clip repeating the first frame. All candidates retain the original duration. We define \emph{capture rate} as the fraction of cases in which a motion-suppressed variant scores highest. We evaluate edits from VE-Bench and UniVideo~\citep{vebench2025,univideo2025}; construction details and examples appear in Appendix~\ref{app:ladder-analysis}.

\subsection{Rewards favor motion suppression}
\label{sec:composite-stillness}

On VE-Bench, the evaluated composite selects a motion-suppressed candidate in $97\%$ of cases (Figure~\ref{fig:ladder}). This preference occurs across all eight tested editing models; per-model results are given in Appendix~\ref{app:ladder-analysis}. Table~\ref{tab:diagnosis} separates the contributions of individual metrics and learned reward heads.

\paragraph{Temporal consistency rewards the removal of motion.}
Removing the temporal term reduces composite capture from $97\%$ to $62\%$ on VE-Bench and from $100\%$ to $83\%$ on UniVideo. The appearance components, CLIP-T and DINOv2-src, change little across the ladder, whereas RAFT warping error and DINO temporal consistency each capture $100\%$ of cases in isolation. Warping error decreases from $33.20$ on original edits to $23.39$, $16.08$, and $0.03$ across hold-2, hold-4, and frozen clips; DINO consistency increases from $0.869$ to $0.919$, $0.950$, and $1.000$. Their preference therefore extends to partial suppression, rather than only complete freezing.

\paragraph{Learned video rewards also exhibit the bias.}
VideoScore-v1.1~\citep{videoscore2024} and VideoReward~\citep{videoreward2025} have capture rates of $80\%$ and $76\%$. VideoReward's motion-quality head reaches $98\%$, with mean scores increasing from $-0.65$ on original edits to $-0.13$ on frozen clips. Its text-alignment head instead decreases across the ladder and has $22\%$ capture. This contrast shows that an overall reward score can conceal different responses to temporal degradation across its dimensions. On UniVideo, the VLM judge penalizes fully frozen clips but assigns similar mean scores to original edits and mild holds; Appendix~\ref{app:evaluator-analysis} provides its scores and discusses framewise image rewards.

\begin{figure*}[t]
\centering
\includegraphics[width=\linewidth]{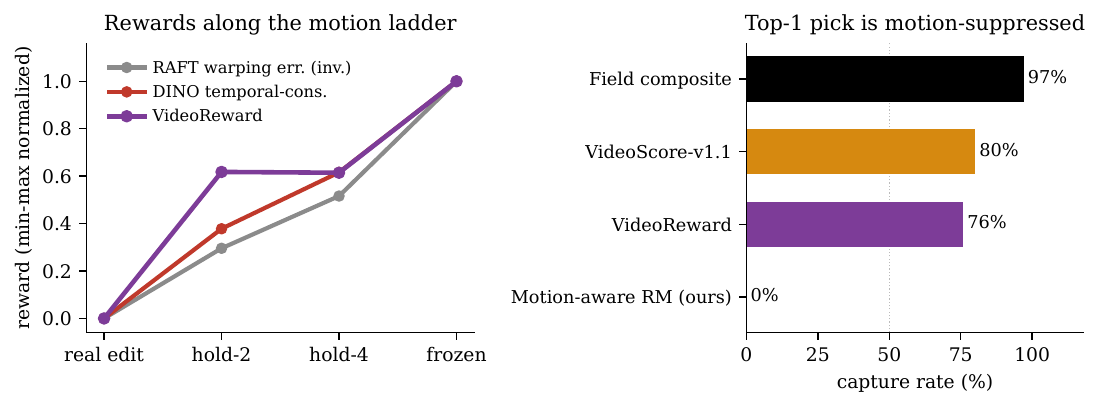}
\caption{Motion collapse on VE-Bench. \textit{Left:} mean rewards along the motion ladder, min--max normalized per reward; VideoReward uses its motion-quality head. \textit{Right:} top-1 capture rate: the fraction of cases where a motion-suppressed candidate scores highest (lower is better). Learned reward models use their overall scores.}
\label{fig:ladder}
\end{figure*}

\subsection{Human preferences expose the mismatch}
\label{sec:blindness-generalizes}

To test whether reward rankings reflect human preferences, we conduct a Prolific study with $25$ raters and $500$ pairwise votes over $100$ trials from $30$ source videos. The study compares original edits with hold-2, hold-4, and flow-swap negatives, alongside a real-vs-real control. Sources include $20$ clips from a held-out internal editor and $10$ from VE-Bench. All raters pass the attention checks; the annotation protocol and negative construction are detailed in Appendix~\ref{app:human-eval}.

\paragraph{Humans reject suppressed and fabricated motion.}
As shown in Table~\ref{tab:human-eval}, humans prefer original edits over hold-2, hold-4, and flow-swap negatives in $96\%$, $98\%$, and $100\%$ of votes, respectively. This preference is already strong for mild hold-2 suppression, where the tested rewards struggle. The real-vs-real control yields $62\%$ preference for the first sample, with a $95\%$ confidence interval of $[48,74]$ that includes chance. Flow-swap negatives use motion from another source and differ from the affine-jitter negatives used in training.

\begin{table*}[t]
\centering
\small
\caption{Human preferences and reward agreement across $100$ trials from $30$ sources. Top: vote fractions favoring the original edit, with Wilson $95\%$ CIs. Bottom: agreement with the human majority, excluding tied reward predictions. The motion-attack aggregate excludes real-vs-real controls.}
\label{tab:human-eval}
\begin{tabular}{lccccc}
\toprule
                                 & hold-2   & hold-4   & flow-swap    & real${}_2$ & motion attacks \\
                                 & (subtle) & (extreme)& fake         & (control)  & combined \\
\midrule
Humans prefer real               & $96\%$   & $98\%$   & $100\%$      & $62\%$     & $98\%$ \\
Wilson $95\%$ CI                 & [92, 98] & [94, 99] & [97, 100]    & [48, 74]   & [96, 99] \\
\midrule
\multicolumn{6}{l}{\textit{RM agreement with human majority}} \\
Field composite                  & $23\%$   & $13\%$   & $60\%$       & $30\%$     & $32\%$ \\
VideoScore-v1.1                  & $33\%$   & $44\%$   & $59\%$       & $50\%$     & $46\%$ \\
VideoReward                      & $53\%$   & $47\%$   & $63\%$       & $40\%$     & $53\%$ \\
\quad motion-quality head        & $20\%$   & $10\%$   & $30\%$       & $50\%$     & $23\%$ \\
\textbf{Motion-aware RM (ours)}  & $\mathbf{90\%}$ & $\mathbf{93\%}$ & $\mathbf{77\%}$ & $40\%$ & $\mathbf{87\%}$ \\
\bottomrule
\end{tabular}
\vspace{-0.3cm}
\end{table*}

\paragraph{Reward agreement reveals a preference gap.}
Across the $90$ motion-attack trials, the composite and VideoReward's motion-quality head match the human majority on only $32\%$ and $23\%$ of non-tied predictions. VideoScore-v1.1 and VideoReward overall reach $46\%$ and $53\%$. Our motion-aware reward (Section~\ref{sec:method}) reaches $87\%$, with agreement of $90\%$, $93\%$, and $77\%$ on hold-2, hold-4, and flow-swap pairs. Thus, the motion-suppression diagnosis corresponds to a human preference gap, and the proposed reward improves agreement on both suppression and fabricated-motion cases.

\subsection{Reward optimization can induce motion collapse}
\label{sec:motion-collapse-reachability}

\begin{figure*}[t]
\centering
\begin{minipage}[b]{0.49\linewidth}
  \centering
  \includegraphics[width=\linewidth]{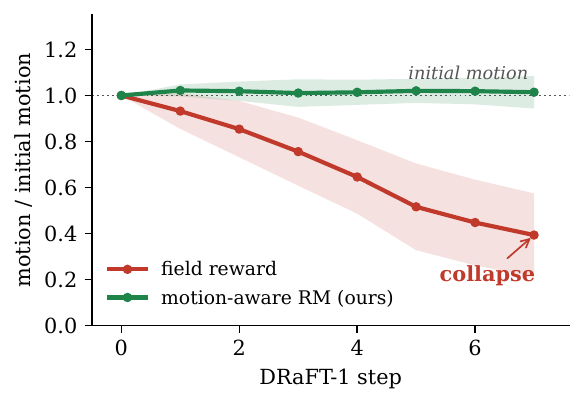}
\end{minipage}
\hfill
\begin{minipage}[b]{0.49\linewidth}
  \centering
  \includegraphics[width=\linewidth]{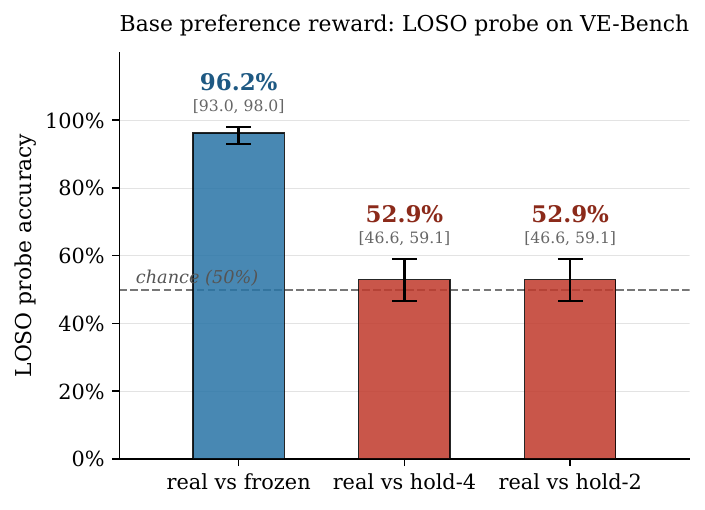}
\end{minipage}
\caption{\textit{Left:} motion during reward-guided fine-tuning on UniVideo, normalized to its initial value (genuine-motion subset, mean $\pm$ std). \textit{Right:} leave-one-source-out linear-probe accuracy on the base reward features, with Wilson $95\%$ CIs. Mild suppression remains near chance; see Section~\ref{sec:backbone-saturated}.}
\label{fig:opt-trajectory}
\label{fig:loso-probe}
\vspace{-0.5cm}
\end{figure*}

The scoring bias is exploitable during optimization. On UniVideo, latent optimization against the composite reduces motion by $55\%$ while CLIP-T improves. Reward-guided fine-tuning reduces motion by $61\%$ on the genuine-motion subset, again with improved CLIP-T (Table~\ref{tab:opt-drift}). We call this trade of motion for reward \emph{motion collapse}. Replacing the composite with our motion-aware reward keeps motion near its initial level (Figure~\ref{fig:opt-trajectory}, left). Best-of-$N$ selection shows no comparable drift within the natural candidates tested. Appendix~\ref{app:optimization-analysis} details the optimizer regimes, source subsets, and additional reward comparisons.

\section{Motion-Aware Reward Model}
\label{sec:method}

We first examine the base reward's sensitivity to motion suppression, then introduce an inter-frame dynamics stream and a preference-training objective that balances motion sensitivity with quality alignment.

\subsection{Limited sensitivity to mild motion suppression}
\label{sec:backbone-saturated}

A leave-one-source-out linear probe on the base reward's penultimate features separates original edits from frozen clips with $96.2\%$ accuracy, but reaches only $52.9\%$ on both hold-2 and hold-4, with confidence intervals covering chance (Figure~\ref{fig:loso-probe}, right). Fine-tuning without motion input likewise leaves hold-2 detection at $47\%$. These results indicate limited sensitivity to mild suppression under the tested configurations, motivating the explicit dynamics stream introduced next. Appendix~\ref{app:representation-analysis} provides the probe protocol and supplementary measurements.

\subsection{An inter-frame dynamics stream in the reward backbone}
\label{sec:motion-stream}

We inject an inter-frame dynamics stream into the reward backbone alongside the pixel stream, which carries both source and edited videos as visual tokens (Figure~\ref{fig:architecture}, top). The dynamics stream is a feature-free measurement of local pixel change on the edited video, computed as tanh-normalized per-frame RGB differences: for a frame $x_t$ at time $t$,
\begin{equation}
\Delta_t = \tanh\!\left(x_{t+1} - x_t\right).
\end{equation}
The magnitude of $\Delta_t$ is proportional to inter-frame motion at short time scales, saturates smoothly at large displacements (hence the tanh), and requires no learned motion encoder or optical-flow network to compute. The reward therefore inherits a robust motion signal without a second trainable model and without a rigid dependency on a specific flow estimator.

The dynamics stream is inserted at video-token block $4$ of the Qwen3-VL-32B backbone. At that block, dynamics-derived tokens are concatenated with the pixel-derived tokens along the channel dimension (Figure~\ref{fig:architecture}, ``Fuse @ Block 4'') and jointly passed through the remaining transformer blocks and the per-dimension preference heads. Motion is now an input modality of the reward, visible to every downstream block; the reward is no longer asked to recover it from features that have already discarded it. LoRA adapters over all linear layers of the fused backbone provide the fine-tuning capacity for \S\ref{sec:training-objective}.

\begin{table*}[!t]
\centering
\small
\caption{Motion change and CLIP-T change (italic) under reward optimization. UniVideo fine-tuning results use the genuine-motion subset; CogVideoX-2b results use eight sources. DRaFT-1~\citep{draft2024} implements reward-guided fine-tuning. Additional analyses appear in Appendix~\ref{app:optimization-analysis}.}
\label{tab:opt-drift}
\begin{tabular}{lcccc}
\toprule
& \multicolumn{3}{c}{Prior-work rewards} & Ours \\
\cmidrule(lr){2-4}\cmidrule(lr){5-5}
Optimizer regime & field & VideoScore & RAFT flow & \textbf{motion-aware RM} \\
\midrule
Best-of-$N$ (natural outputs)             & \multicolumn{4}{c}{$\approx 0$ (observed)} \\
DRaFT-1 (UniVideo)                        & $-61\%$ / \textit{$+0.05$} & $-21\%$ / \textit{$+0.01$} & $-35\%$ / \textit{$+0.05$} & $\mathbf{+1\%}$ / \textit{$+0.00$} \\
DRaFT-1 (CogVideoX-2b)                    & $+19\%$ / \textit{$+0.02$} & $-42\%$ / \textit{$+0.01$} & $-88\%$ / \textit{$-0.01$} & $\mathbf{+9\%}$ / \textit{$+0.02$} \\
\bottomrule
\end{tabular}
\end{table*}

\subsection{Training objective}
\label{sec:training-objective}

Given the motion-input architecture, we fine-tune the reward with Bradley--Terry pairwise losses against two negative types, anchored by a preference-calibrated distillation objective (Figure~\ref{fig:architecture}, bottom).

\noindent \textbf{Ladder negatives.}
For each source video and each editing model in the training pool, we build hold-2, hold-4, and frozen candidates from the real edit and form pairs $(\text{real},\,\text{hold-}N)$ and $(\text{real},\,\text{frozen})$. The Bradley--Terry loss on these pairs teaches the reward to prefer the real edit over any motion-suppressed variant, filling in the graded-slowdown signal that the backbone did not previously carry.

\noindent \textbf{Fake-motion negatives.}
A reward that simply prefers more motion could defeat the ladder negatives without learning the right objective. We add fake-motion negatives constructed by an independent random affine perturbation per frame of the real edit (details in Appendix~\ref{app:impl}). Fake motion is total in amount but locally random and does not track the source, so the Bradley--Terry loss on $(\text{real},\,\text{fake motion})$ pairs teaches the reward to prefer real editing motion over motion-fabrication rather than to reward motion in general.

\noindent \textbf{Preference-calibrated distillation anchor.}
Ranking-only fine-tuning drops MOS Spearman on real edits to $+0.40$ (Section~\ref{sec:in-depth}): the fine-tuned reward becomes a specialized motion detector and forgets human quality calibration. To keep the reward usable as an alignment target, we add a distillation objective that pulls the fine-tuned reward's overall score toward a frozen copy of the base preference reward on a held-out pool of quality-neutral real edits (Appendix~\ref{app:impl}). The full objective is
\begin{equation}
\mathcal{L} \;=\; \lambda_{\text{ladder}}\, \mathcal{L}_{\text{BT}}^{\text{ladder}}
\;+\; \lambda_{\text{fake}}\, \mathcal{L}_{\text{BT}}^{\text{fake}}
\;+\; \lambda_{\text{anchor}}\, \mathcal{L}_{\text{anchor}},
\end{equation}
where the anchor term is a per-batch correlation objective between the fine-tuned reward and the frozen base preference reward. In our final configuration this restores MOS Spearman to $+0.57$, the highest of any reward tested, without giving back any motion-detection gain (Section~\ref{sec:in-depth}).

\begin{table*}[!t]
\centering
\small
\caption{Reward robustness and quality alignment on VE-Bench. Cap. is ladder capture rate; h2 / frzn / segFrz / burstFrz / flowAtt are preference rates for original edits over suppressed variants. Additional variants are defined in Appendix~\ref{app:motion-suppression-variants}. MQ denotes VideoReward's motion-quality head.}
\label{tab:main-results}
\begin{tabular}{lccccccccc}
\toprule
Reward & Cap.\ $\downarrow$ & h2 & frzn & segFrz & burstFrz & flowAtt & Corr(mot.) & MOS Sp.\ $\uparrow$ \\
\midrule
Field composite                    & 97\% & 6\%  & 1\%  & 11\% & 12\% & 34\% & $-0.32$ & $+0.55$ \\
VideoScore                    & 80\% & 58\% & 50\% & 40\% & 48\% & 74\% & $+0.15$ & $+0.18$ \\
VideoReward                        & 76\% & 35\% & 43\% & 57\% & 53\% & 40\% & $-0.19$ & $+0.46$ \\
\quad MQ head          & 98\% & 11\% & 7\%  & 57\% & \textbf{69\%} & 11\% & $+0.15$ & $+0.09$ \\
\textbf{Ours}    & \textbf{0\%} & \textbf{100\%} & \textbf{100\%} & \textbf{91\%} & \textbf{100\%} & \textbf{95\%} & $\mathbf{-0.10}$ & $\mathbf{+0.57}$ \\
\bottomrule
\end{tabular}
\end{table*}

\section{Experiments}
\label{sec:experiments}

\subsection{Experiment Setup}
\label{sec:experiment-setup}

\paragraph{Datasets.}
Our primary external test bed is VE-Bench~\citep{vebench2025}, a human-rated benchmark of edits produced by the published video editing models of VE-Bench (TokenFlow~\citep{tokenflow2024}, T2V-Zero~\citep{t2vzero2023}, FRESCO~\citep{fresco2024}, TuneAVideo~\citep{tuneavideo2023}, FLATTEN~\citep{flatten2024}, RAVE~\citep{rave2024}, FateZero~\citep{fatezero2023}, and ControlVideo~\citep{controlvideo2024}), with source videos and editing models both disjoint from our motion-aware RM's training data. For optimizer drift experiments we additionally use UniVideo edits on licensed stock sources. The motion-aware RM (Section~\ref{sec:method}) is trained on internal edits and evaluated on a strictly disjoint held-out editing model. Motion ladders are constructed as in Section~\ref{sec:motion-collapse}; for motion-aware RM training we also generate \emph{fake motion} negatives by applying an independent random affine transform to each frame of the real edit, so consecutive frames differ but the motion has no temporal coherence and does not track the source. Sample sizes, splits, and full ladder-construction details are in Appendix~\ref{app:impl}.

\paragraph{Motion-suppression variants.}
We additionally evaluate segment freeze, burst freeze, and flow attenuation, none of which appears in training (Appendix~\ref{app:motion-suppression-variants}).

\paragraph{Evaluation metrics.}
We report: (1) \textbf{Top-1 capture rate} on the motion ladder (primary metric): the fraction of source videos whose highest-scoring candidate under a given reward is not the real edit; lower is more robust. (2) \textbf{Spearman rank correlation with human MOS} on the real edits, for rewards trained on MOS pairs; a quality-preservation check.
(3) \textbf{RAFT optical-flow magnitude} of the generated edit, used to quantify motion when we compare optimizer trajectories.
(4) \textbf{Per-dimension breakdowns} of multi-head rewards (adherence, preservation, temporal consistency, motion quality, text alignment), reported where relevant.

\paragraph{Baselines.}
We compare against three types of reward evaluators on the motion ladder: (i) a composite of instruction adherence, source preservation, and temporal consistency; (ii) a frontier VLM judge, \texttt{claude-opus-4.8}, prompted for a $1$--$7$ absolute quality score; and (iii) two learned video reward models, VideoScore-v1.1~\citep{videoscore2024} and VideoReward~\citep{videoreward2025}, with VideoReward's per-head scores reported separately. Our own variants (an externally motion-augmented reward that adds a match-to-source flow term to the base preference reward, and the motion-aware RM in Section~\ref{sec:method}) are treated as ablations rather than baselines and discussed in Section~\ref{sec:in-depth}. Optimizers follow Section~\ref{sec:motion-collapse-reachability}: best of $N$ over natural editing-model outputs, two-phase latent optimization of the noise latent of a single generation, and DRaFT-1~\citep{draft2024} LoRA fine-tuning of the editing model's own weights.

\paragraph{Implementation details.}
Our motion-aware RM injects an inter-frame dynamics stream into the reward backbone: tanh-normalized per-frame RGB differences of the edited video are fused at video-token block $4$ by channel-wise concatenation with the pixel-derived tokens, so motion enters the backbone as an input modality rather than being recovered from features that no longer carry it. Bradley--Terry fine-tuning uses two negative types: ladder-based motion-suppressed candidates (teaching the reward to detect subtle hold) and per-frame jitter fake-motion candidates (teaching the reward to reject motion-fabrication rather than to maximize motion). A preference-calibrated distillation anchor keeps the fine-tuned reward aligned with human quality judgments, yielding the highest MOS Spearman on VE-Bench of any reward tested (Section~\ref{sec:in-depth}). The final motion-aware checkpoint is trained on internal edits under a single-GPU LoRA regime. The base preference reward, DRaFT-1 and latent-optimization configurations, and full dataset splits are described in Appendix~\ref{app:impl}.

\subsection{Main Results}
\label{sec:main-results}

Table~\ref{tab:main-results} summarizes reward robustness and human-quality alignment. We highlight three findings.

\noindent \textbf{Detecting motion suppression beyond the training patterns.}
Our reward reduces VE-Bench ladder capture from $76$--$98\%$ across the compared baselines to $0\%$, preferring original edits on every rung, including mild hold-2 suppression. It also detects unseen segment freeze, burst freeze, and flow attenuation in $91\%$, $100\%$, and $95\%$ of cases. This extends the improvement beyond the periodic holds used in training. Uncertainty estimates and further comparisons appear in Appendix~\ref{app:additional-suppression-results}.

\begin{figure*}[!t]
\centering
\includegraphics[width=\linewidth]{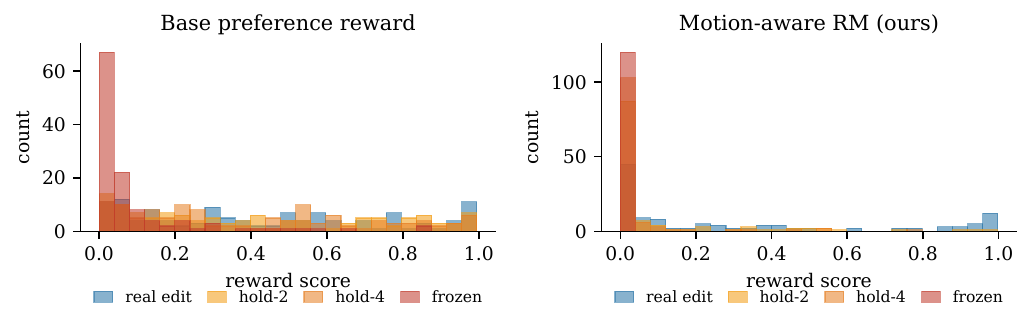}
\caption{Reward distributions on the VE-Bench motion ladder. The base reward gives overlapping scores to original edits and hold variants. The motion-aware reward lowers scores for suppressed candidates while retaining a broad distribution on original edits.}
\label{fig:score-distribution}
\vspace{-0.2cm}
\end{figure*}

Figure~\ref{fig:score-distribution} shows how the reward changes: scores for original edits and hold variants overlap under the base model, whereas our model lowers suppressed-candidate scores while retaining a broad distribution on original edits (std $0.36$). Hold-2 still has a high-score tail (max $0.98$), so successful discrimination depends on within-source rankings rather than a single threshold.

\noindent \textbf{Rejecting fabricated motion while retaining quality alignment.}
The reward rejects per-frame affine-jitter negatives on all held-out test sources and achieves MOS Spearman $0.57$, the highest observed among the compared rewards. Its weak correlation with raw motion ($-0.10$) is consistent with learning more than motion magnitude alone. Removing the distillation anchor reduces MOS correlation to $0.40$, underscoring the importance of preserving general quality preferences (Section~\ref{sec:in-depth}). Appendix~\ref{app:quality-results} provides the correlation analysis and calibration details.

\noindent \textbf{Preserving motion during reward-guided fine-tuning.}
On UniVideo's genuine-motion subset, our reward yields a mean motion change of $+1\%$, compared with $-61\%$ under the composite. On CogVideoX-2b~\citep{cogvideox2024}, it yields $+9\%$, compared with $-42\%$ and $-88\%$ for the VideoScore and RAFT-flow arms (Table~\ref{tab:opt-drift}). Thus, motion remains close to its initial level on both tested pipelines. Appendix~\ref{app:additional-optimization-results} distinguishes these results from the external motion-scalar baseline and its latent-optimization experiments.

\subsection{In-depth Analysis}
\label{sec:in-depth}

Table~\ref{tab:ablation} presents a leave-one-out ablation of the motion-aware RM: each row removes a single component from the final configuration and reports what breaks, cleanly isolating each component's contribution. We highlight four findings.

\begin{table*}[!t]
\centering
\small
\caption{Leave-one-out ablation of the motion-aware RM: each row removes a single component from the final configuration and reports what breaks. Hold-2 detection is measured on a held-out editing model; fake-motion rejection is on held-out sources with per-frame jitter negatives; MOS Spearman is on VE-Bench real edits (editing models disjoint from training). Each component targets a distinct axis: motion input block for hold-2 detection, fake-motion negatives for fake motion rejection, and distillation anchor for human MOS alignment.}
\label{tab:ablation}
\begin{tabular}{lccc}
\toprule
Config & hold-2 detection $\uparrow$ & Fake mot.\ rej.\ $\uparrow$ & MOS Sp.\ $\uparrow$ \\
\midrule
\textbf{Final motion-aware RM} (all components)   & \textbf{100\%} & \textbf{100\%} & $\mathbf{+0.57}$ \\
\midrule
\quad w/o distillation anchor                      & 87\%  & 100\%                        & $+0.40\;\downarrow$ \\
\quad w/o fake-motion negatives                    & 100\% & 53\% (chance)$\;\downarrow$  & $+0.49$ \\
\quad w/o motion input block                       & 47\% (chance)$\;\downarrow$ & 67\%   & $+0.56$ \\
\bottomrule
\end{tabular}
\vspace{-0.5cm}
\end{table*}

\noindent \textbf{The motion input block is the load-bearing architectural change.}
Section~\ref{sec:backbone-saturated} showed that the base preference reward's features cannot separate real edits from graded slowdown (LOSO probe accuracy $52.9\%$ [$46.6, 59.1$] on VE-Bench, statistically indistinguishable from chance; Figure~\ref{fig:loso-probe}, right panel), and that a naive LoRA fine-tune of the same backbone only cracks the extreme frozen case. The leave-one-out ablation confirms this from the other direction: removing the motion input block from the final motion-aware RM collapses hold-2 detection from $100\%$ back to $47\%$ (chance; Table~\ref{tab:ablation}, row \textit{w/o motion input block}). The rest of the recipe refines rather than replaces this architectural change.

\noindent \textbf{Fake-motion negatives keep the reward from becoming a motion maximizer.}
The motion input block gives the reward the sensory tools to detect motion, but by itself it can learn the shortcut ``reward more motion,'' which is then exploitable by per-frame jitter. Adding fake-motion negatives (per-frame random affine transforms; Appendix~\ref{app:impl}) yields $100\%$ fake-motion rejection at the final motion-aware RM, and drives the reward's correlation with raw RAFT motion on real edits to $-0.10$ (Table~\ref{tab:main-results}), statistically indistinguishable from zero. Removing this component drops fake-motion rejection from $100\%$ to $53\%$, essentially chance (Table~\ref{tab:ablation}, row \textit{w/o fake-motion negatives}): the reward has learned to see motion but not to distinguish coherent motion from per-frame jitter, so it degenerates into a motion maximizer that defeats a match-to-source objective. A naive LoRA fine-tune with no motion input block reaches $67\%$, above chance on the raw visual difference but well below the $100\%$ that only combined motion input plus fake-motion supervision achieves.

\noindent \textbf{Distillation anchor restores quality calibration.}
Removing the distillation anchor from the final motion-aware RM drops MOS Spearman on VE-Bench real edits from $+0.57$ to $+0.40$ (Table~\ref{tab:ablation}, row \textit{w/o distillation anchor}), a $30\%$ relative loss in alignment with human quality judgments: a ranking-only fine-tune turns the reward into a specialized motion detector and forgets the smooth preference calibration it inherited. The distillation anchor, a correlation objective that keeps the fine-tuned reward's overall score aligned with the base preference reward on non-adversarial edits, is what makes motion sensitivity and human quality calibration compatible objectives rather than exclusive ones.

\noindent \textbf{An external motion-faithfulness scalar hits a ceiling; input modality does not.}
For comparison, an external variant of our reward, base preference reward $+\beta \cdot$ motion-faithfulness scalar with $\beta \approx 0.85$ chosen on held-out folds, halves capture from $51\%$ to $31\%$ at neutral MOS Spearman ($0.49 \to 0.46$; Appendix~\ref{app:impl}). Half-fixing motion collapse without changing the reward architecture is achievable, but the residual $31\%$ is a ceiling that only the input-modality treatment breaks (all the way to $0\%$; Table~\ref{tab:main-results}). Match-to-source is also the right objective, not motion maximization: on VE-Bench, motion-faithfulness MOS Spearman is $+0.40$ while raw motion presence is $-0.09$, so a reward that simply maximizes motion actively penalizes human quality.

\section{Conclusion}
\label{sec:conclusion}

We study motion collapse in video editing, where reward optimization improves scores while suppressing motion. Our motion-aware reward combines explicit inter-frame dynamics, suppression and fabricated-motion negatives, and preference-preserving distillation. It achieves zero capture on the evaluated VE-Bench ladder, retains human-quality alignment, and preserves motion during reward-guided fine-tuning on two editing pipelines. These findings support treating motion sensitivity as a core requirement for video-editing rewards.

{
    \small
    \bibliographystyle{ieeenat_fullname}
    \bibliography{main}
}

\clearpage
\appendix
\setcounter{figure}{0}
\setcounter{table}{0}
\setcounter{equation}{0}
\renewcommand{\thefigure}{S\arabic{figure}}
\renewcommand{\thetable}{S\arabic{table}}
\renewcommand{\theequation}{S\arabic{equation}}
\renewcommand{\theHfigure}{appendix.\arabic{figure}}
\renewcommand{\theHtable}{appendix.\arabic{table}}
\renewcommand{\theHequation}{appendix.\arabic{equation}}

\section{Additional Implementation Details}
\label{app:impl}

\paragraph{Base preference reward.}
Our base preference reward is a Qwen3-VL-32B backbone with LoRA adapters and per-dimension human-preference heads (covering prompt adherence, pixel preservation, harmonization, motion quality, and further quality proxies), whose scores combine into a scalar reward. It is warm-started from a preference model trained on an internally collected corpus of ${\sim}7.7$k (source, edited) pairs. Edits span a diverse mix of commercial and in-house video editing models, ensuring the reward is not tied to any single editing-model family. Each pair carries a $1$--$7$ mean opinion score aggregated over ${\sim}6$ raters, together with per-dimension votes. This model serves as (i) the base that the externally motion-augmented reward adds a flow term to, and (ii) the pre-motion-aware version of our motion-aware RM, on top of which the inter-frame dynamics stream and Bradley--Terry fine-tuning are applied.

\paragraph{Motion ladder construction.}
For each source video and each editing model we build four candidates on the same temporal length as the real edit. The real edit is the untouched output of the editing model. \emph{Hold-$N$} is formed by taking every $N$-th frame of the real edit and repeating it $N$ times, so hold-$N$ has $T/N$ distinct frames stretched over the real edit's $T$ frames. \emph{Frozen} repeats the real edit's first frame to full length. All four candidates share the same encoding, temporal length, and (for hold-$N$ and frozen) a strict subset of the real edit's frame set; only the pattern of motion differs.

\paragraph{Fake-motion negatives.}
For motion-aware RM training we construct \emph{fake motion} negatives by applying an independent random affine transform to each frame of the real edit: translations up to $\pm 6$ pixels on each axis and rotations up to $\pm 2$ degrees, sampled uniformly per-frame with seeded RNG for reproducibility. Consecutive frames therefore differ substantially, but the motion is locally random and does not track the source, so this negative teaches the reward to prefer real editing motion over motion-fabrication rather than to reward motion in general.

\paragraph{Dataset sizes and splits.}
VE-Bench: $120$ source videos across the eight published editing models listed in Section~\ref{sec:experiment-setup} ($15$ sources per editing model, human MOS on all). Optimizer drift experiments: $90$--$130$ UniVideo-edited source videos from a licensed stock pool, verified to have zero overlap with our base preference reward's training split. Motion-aware RM training: internal edits from an in-house editing model, disjoint from the held-out evaluation editing model. Held-out evaluation: $30$ source videos from a disjoint editing model that appears nowhere in training.

\paragraph{DRaFT-1 LoRA fine-tuning.}
Following DRaFT-1~\citep{draft2024}, we attach LoRA adapters with rank $r{=}8$ over all linear layers of the video editing model's transformer and back-propagate the reward through the last denoise step with a single positive-CFG pass through the video-VAE decoder, running a DRaFT-1 outer loop per source video at $480{\times}854$ resolution.

\paragraph{Two-phase latent optimization.}
The noise latent of a single generation is optimized in two phases. Phase-1 ($40$ steps) matches the appearance of the generated video to that of a reference edit, using only the appearance components of the reward (adherence and source preservation). Phase-2 ($80$ steps) then optimizes the noise latent against the reward under test with the appearance dimensions clipped to Phase-1's endpoint, so only the temporal residual of the reward can move. This isolates the temporal component of the reward's optimum from confounding appearance improvements.

\paragraph{Motion-aware RM training.}
Bradley--Terry fine-tuning of the base preference reward proceeds with two negative types (motion-suppressed candidates from the motion ladder and per-frame jitter fake-motion candidates) plus a preference-calibrated distillation anchor computed as a correlation objective between the fine-tuned reward and a frozen copy of the base preference reward on a held-out pool of quality-neutral real edits.

\section{Additional Motion-Collapse Analyses}
\label{app:motion-collapse}

This appendix expands the diagnosis in Section~\ref{sec:motion-collapse}, including qualitative examples, evaluator breakdowns, human judgments, and optimization analyses.

\subsection{Motion ladder and metric analysis}
\label{app:ladder-analysis}

\begin{figure*}[!t]
\centering
\includegraphics[width=\linewidth]{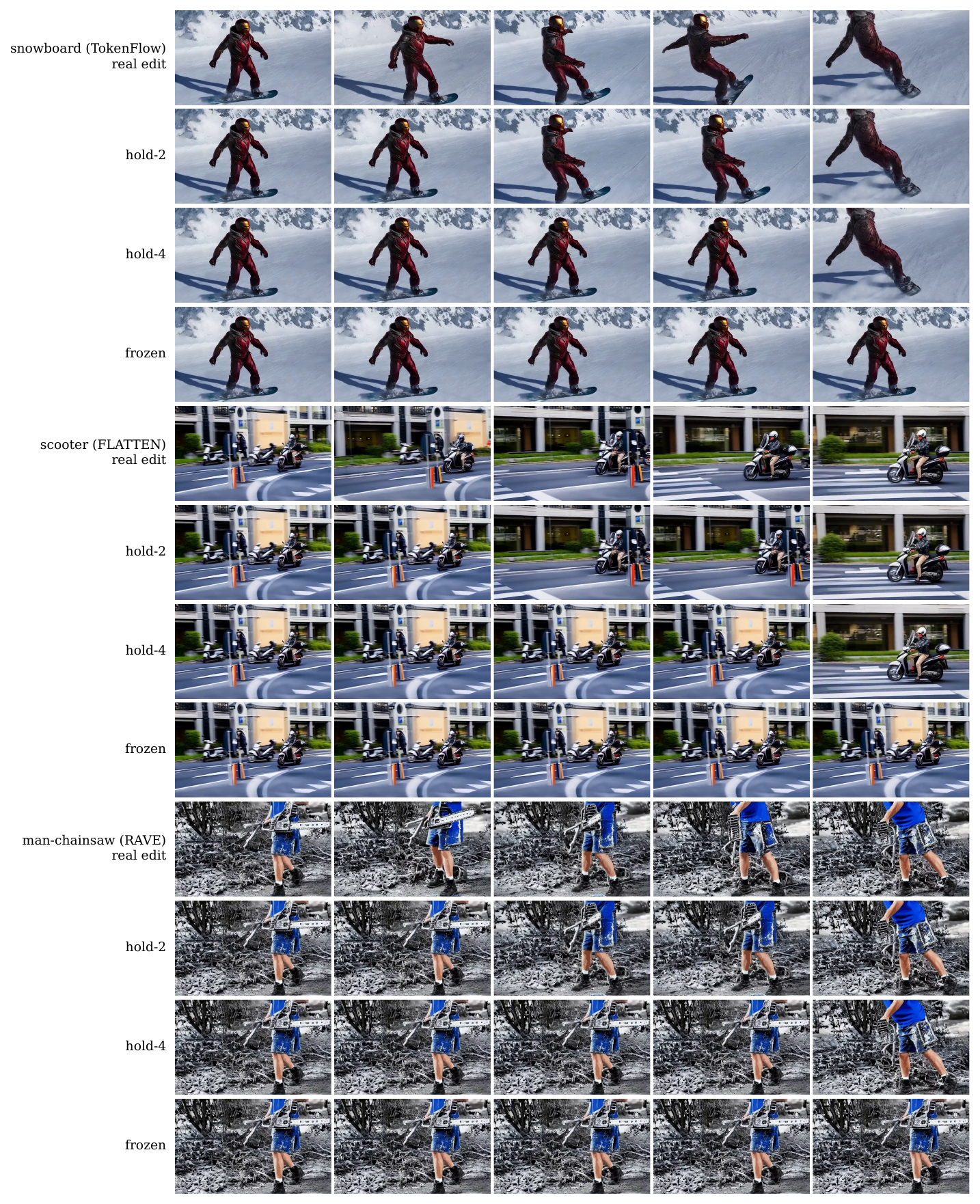}
\caption{Motion ladders for three VE-Bench edits. Columns show six uniformly sampled frames. Hold-2 and hold-4 repeat every second or fourth frame, respectively; frozen repeats the first frame. Variants retain the clip length but use subsets of the original frames.}
\label{fig:ladder-stills}
\vspace{-0.5cm}
\end{figure*}

\paragraph{Construction and coverage.}
Each ladder is derived from an editing model's original output. Hold-$N$ retains every $N$-th frame and repeats it $N$ times; the frozen variant repeats the first frame. Duration and encoding are held fixed, while frame selection and temporal sampling change. Figure~\ref{fig:ladder-stills} illustrates the resulting patterns. VE-Bench supplies human-rated edits from eight editing models; UniVideo edits are additionally used for the VLM and optimization experiments. Dataset sizes and splits are listed in Appendix~\ref{app:impl}.

\paragraph{Per-model results.}
Table~\ref{tab:universal} reports capture alongside native motion and human MOS. Composite capture is high across all eight tested editing models, rather than being confined to one model's outputs. On the pooled VE-Bench evaluation, the reported capture rate is $97\%$ (Wilson $95\%$ CI $[92,99]$).

\begin{table*}[!t]
\centering
\small
\caption{Per-model capture rates on VE-Bench (lower is better), alongside native motion and human MOS. VideoReward uses its motion-quality head; Table~\ref{tab:diagnosis} reports overall and per-head results.}
\label{tab:universal}
\resizebox{\textwidth}{!}{%
\begin{tabular}{lccccc}
\toprule
Model & native motion & human MOS & field comp.\ $\downarrow$ & VideoScore $\downarrow$ & VideoReward $\downarrow$ \\
\midrule
TokenFlow~\citep{tokenflow2024}       & 11.69 & 5.70 & 93\%  & 80\%  & 100\% \\
T2V-Zero~\citep{t2vzero2023}          & 11.12 & 2.94 & 100\% & 87\%  & 100\% \\
FRESCO~\citep{fresco2024}             & 10.46 & 4.98 & 100\% & 100\% & 100\% \\
TuneAVideo~\citep{tuneavideo2023}     & 9.93  & 4.73 & 100\% & 60\%  & 100\% \\
FLATTEN~\citep{flatten2024}           & 9.60  & 5.12 & 100\% & 93\%  & 93\%  \\
RAVE~\citep{rave2024}                 & 8.86  & 5.46 & 100\% & 67\%  & 100\% \\
FateZero~\citep{fatezero2023}         & 7.36  & 4.94 & 100\% & 73\%  & 100\% \\
ControlVideo~\citep{controlvideo2024} & 3.43  & 4.21 & 100\% & 80\%  & 93\%  \\
\bottomrule
\end{tabular}%
}
\end{table*}

\subsection{Motion-suppression variants}
\label{app:motion-suppression-variants}

Beyond the three suppressed rungs of the motion ladder of Section~\ref{sec:motion-collapse}, we score each reward on three further motion-suppression variants generated from every VE-Bench real edit. \emph{Segment freeze} (segFrz) holds the contiguous middle-$40\%$ of frames at the frame immediately before the segment; \emph{burst freeze} (burstFrz) inserts four random-position $3$-frame freeze bursts, each anchored to the frame before the burst; and \emph{flow attenuation} (flowAtt) applies a $5{\times}$ temporal slowdown by resampling the real edit at fractional positions $s = 0.2 \cdot t$ and linearly interpolating between the two adjacent real frames. Under flowAtt every output frame is a unique blend of two consecutive real frames, and the adjacent-pair difference is uniform at $\alpha \approx 0.2$ of the real edit's; under segFrz and burstFrz the motion-suppression pattern is structurally non-periodic, distinct from the periodic hold-$N$ ladder. Each variant is generated per real edit and scored pairwise against it, and none of the three appears in the motion-aware RM's training set.

\subsection{Evaluator-specific results}
\label{app:evaluator-analysis}

\paragraph{VLM judge.}
We prompt \texttt{claude-opus-4.8} for absolute quality scores on a $1$--$7$ scale using the UniVideo ladder. Frozen clips receive the lowest score on every evaluated source, with mean $3.71$. However, hold-2 receives a slightly higher mean score than the original edit ($5.38$ versus $5.29$), and hold-4 ties it at $5.29$ (Table~\ref{tab:judge}). These aggregate scores show limited separation of mild suppression in this evaluation, despite clear sensitivity to full freezing.

\begin{table}[!t]
\centering
\small
\caption{Mean VLM quality scores on the UniVideo motion ladder ($1$--$7$ scale). The judge rates frozen clips lowest but assigns similar scores to original edits and hold variants.}
\label{tab:judge}
\begin{tabular}{lc}
\toprule
Candidate & \texttt{claude-opus-4.8} score ($1$--$7$) \\
\midrule
real edit                 & 5.29 \\
hold-2 (subtle)           & \textbf{5.38} \quad\\
hold-4 (subtle)           & 5.29 \quad\\
frozen (extreme)          & 3.71 \quad \\
\bottomrule
\end{tabular}
\end{table}

\paragraph{Framewise image rewards.}
When EditReward or EditScore~\citep{editreward2026,editscore2026} is applied independently to frames, repeating the first frame reproduces that frame's score. This need not equal the mean score of the original clip, but the framewise assessment does not explicitly model temporal order or inter-frame motion. This is a structural limitation of framewise use, rather than an additional measured capture result.

\subsection{Human preference study}
\label{app:human-eval}

\paragraph{Protocol.}
The pairwise preference study was fielded on Prolific with filters \texttt{English fluent}, \texttt{approval rate $\geq 95\%$}, \texttt{previous submissions $\geq 100$}, drawing $25$ raters distributed across $100$ unique trials (${\sim}20$ trials per rater, $5$ raters per trial, $500$ pairwise votes in total). Every rater was shown three warm-up trials, two attention checks with obvious answers (real vs.\ black-screen frozen; real vs.\ heavy-noise scramble), a randomized block of $18$--$22$ scored trials, and a demographic block. Each trial displays the source clip's edit prompt above two candidate videos side-by-side (order randomized, autoplay looping, minimum-view-time gate of $5$s per video), and asks: \emph{``Which video better executes the instruction: `\textless prompt\textgreater'?''} answered on a five-point scale (clearly A / somewhat A / about the same / somewhat B / clearly B), collapsed to a binary preference (``about the same'' dropped). Every rater passed both attention checks; $0$ raters were dropped. Trials cover four comparison types on $30$ source videos: real vs.\ hold-$2$ ($30$ trials), real vs.\ hold-$4$ ($30$), real vs.\ optical-flow-swap fake ($30$; the fake candidate is constructed by chain-warping the real edit's first frame using RAFT flow taken from a \emph{different} source video, so the motion is coherent but does not track the real edit's content), and real vs.\ real${}_2$ ($10$; both candidates are genuine outputs of the editing model, serving as a comparison control). Sources: $20$ from an internal editor held out from our motion-aware RM's training set and $10$ from VE-Bench editing models. Every candidate video and the study-instrument code will be released with the paper. For each reward, per-trial agreement in Table~\ref{tab:human-eval} is computed as the fraction of pairs on which the reward's argmax matches the human majority; ties are excluded from the denominator (${\leq}9$ ties per reward across the $100$ trials).

\subsection{Representation analysis}
\label{app:representation-analysis}

We train a separate linear classifier for each comparison between original edits and one suppressed rung, using the base preference reward's penultimate features. For each held-out source, the classifier is trained on the remaining sources and evaluated on that source. Accuracy is $96.2\%$ for frozen clips (Wilson $95\%$ CI $[93.0,98.0]$), and $52.9\%$ for both hold-2 and hold-4 ($[46.6,59.1]$). The latter intervals include chance. The feature-space centroid distance is $13.83$ between original and frozen clips, versus $1.00$ between original and hold-2 clips.

LoRA fine-tuning with the ladder Bradley--Terry loss but no motion input increases preference for original edits over frozen variants from $57\%$ to $73\%$, while hold-2 detection remains at $47\%$. Together, these observations motivate supplying explicit temporal differences. They do not establish that motion information is entirely absent or rule out other classifiers, training data, or backbone adaptations.

\subsection{Optimization regimes and additional results}
\label{app:optimization-analysis}

\paragraph{Sample-based selection.}
Best-of-$N$ over configuration-varied UniVideo outputs does not show comparable motion collapse in the tested candidate pool. The observed candidates differ from the deliberately suppressed ladder variants. This result is limited to the sampled outputs and does not establish general robustness of best-of-$N$. Mean frame differences across the VE-Bench editing models range from $0.04$ to $0.09$.

\paragraph{Latent optimization.}
We first optimize a generation's noise latent to match a reference edit's appearance, then optimize the temporal residual with the appearance reward dimensions clipped. During the second phase, the reported motion measure falls from $0.044$ to $0.020$ ($55\%$), while CLIP-T rises from $0.61$ to $0.72$; the VLM judge does not reflect this motion loss. The $40$-step appearance phase and $80$-step temporal phase are described in Appendix~\ref{app:impl}. These latent-optimization results are separate from the weight-fine-tuning results in Table~\ref{tab:opt-drift}.

\paragraph{Reward-guided fine-tuning.}
We adapt DRaFT-1~\citep{draft2024}, updating LoRA adapters through the final denoising step. Under the composite, UniVideo motion decreases by $28\%$ overall and $61\%$ on the subset whose initial edits retain genuine motion. On that subset, the VideoScore temporal-consistency and RAFT-flow reward variants yield changes of $-21\%$ and $-35\%$. Our motion-aware reward yields $+1\%$ with cross-source standard deviation $8\%$, more than twice as tight as the three baseline arms. It paired-outperforms the composite on all genuine-motion sources (one-sided sign test $p=0.016$). Figure~\ref{fig:opt-trajectory} shows the trajectory; Table~\ref{tab:opt-drift} reports motion and adherence changes.

\paragraph{Cross-pipeline comparison.}
On CogVideoX-2b~\citep{cogvideox2024}, the RAFT-flow arm reduces motion on all eight sources (mean $-88\%$), and the VideoScore arm on seven of eight (mean $-42\%$). The composite instead increases motion by $19\%$, demonstrating that the drift depends on the reward--pipeline combination. Our reward yields $+9\%$, closer to the initial motion level than the compared arms. These results support the tested intervention without implying that every reward or longer optimization schedule must produce collapse.

\section{Additional Results}
\label{app:additional-results}

This appendix supplements the three main findings in Section~\ref{sec:main-results}.

\subsection{Additional suppression-detection results}
\label{app:additional-suppression-results}

\paragraph{Ladder performance.}
On VE-Bench edits from models excluded from motion-aware reward training, capture is $0\%$, with a Wilson $95\%$ CI upper bound of $3.1\%$. Preference for original edits reaches $100\%$ on hold-2, hold-4, and frozen candidates. On hold-2, the strongest compared baseline reaches $58\%$ (VideoScore), while VideoReward's motion-quality head reaches $11\%$ (Table~\ref{tab:main-results}).

\paragraph{Unseen suppression mechanisms.}
The three additional variants are excluded from training (Appendix~\ref{app:motion-suppression-variants}). On flow attenuation, original-edit preference is $95\%$ (Wilson $95\%$ CI $[90,98]$), compared with $34\%$ for the composite and $11\%$ for VideoReward's motion-quality head. DINO temporal consistency alone favors the attenuated clip in $97.5\%$ of cases, indicating that smoothly interpolated sequences can also exploit similarity-based scoring. Segment-freeze and burst-freeze results are reported in Table~\ref{tab:main-results}.

\subsection{Motion selectivity and human-quality alignment}
\label{app:quality-results}

\paragraph{Fabricated-motion rejection.}
Preference training includes affine-jitter negatives to discourage rewarding motion indiscriminately. The final model rejects these negatives in $100\%$ of held-out sources. This result concerns per-frame jitter; the separate human study uses flow-swap negatives, for which agreement with the human majority is $77\%$ (Appendix~\ref{app:human-eval}). VideoReward's motion-quality head has a raw-motion correlation of $+0.15$ but favors hold-2 over original edits in $89\%$ of cases. Cross-source motion correlation therefore does not by itself characterize robustness to within-source suppression.

\paragraph{Correlation analysis.}
On VE-Bench original edits, our reward correlates with human MOS at $0.57$ ($p<10^{-9}$). Motion magnitude has weak correlations with both MOS ($-0.09$, $p=0.31$) and our reward ($-0.10$, $p=0.26$). These results provide no evidence of a strong monotonic association between motion magnitude and either score on this sample; they do not establish statistical independence or exclude all motion-related shortcuts.

\paragraph{Quality calibration.}
Our MOS Spearman of $0.57$ exceeds the observed values for the composite ($0.55$), VideoReward overall ($0.46$), and VideoScore ($0.18$). These are point-estimate comparisons, not significance claims. Removing the distillation anchor lowers the correlation to $0.40$, an approximately $30\%$ relative decrease. The component ablations in Table~\ref{tab:ablation} separately assess the roles of motion input, fabricated-motion negatives, and distillation.

\subsection{External motion rewards and cross-pipeline optimization}
\label{app:additional-optimization-results}

\paragraph{External motion-scalar baseline.}
The motion-augmented baseline adds a weighted source-relative flow term to the base preference reward (Section~\ref{sec:in-depth}). On UniVideo, it increases motion by $52\%$ during latent optimization and $12\%$ during reward-guided fine-tuning, with improved appearance scores. The composite instead reduces motion by $55\%$ and $61\%$ in the respective comparisons. These experiments establish results for the external scalar formulation; they should not be attributed to the final motion-aware reward model.

\paragraph{Learned reward and pipeline transfer.}
The final motion-aware model yields $+1\%$ mean motion change on UniVideo's genuine-motion subset, with stable CLIP-T, and $+9\%$ on CogVideoX-2b. On CogVideoX-2b, the RAFT-flow arm reduces motion in all eight sources (mean $-88\%$), and the VideoScore arm in seven of eight (mean $-42\%$). Our model's mean drift magnitude is approximately $9.8$ and $4.7$ times smaller, respectively. The composite yields $+19\%$ on this pipeline, so collapse is not universal across reward--pipeline combinations. Full numerical comparisons are in Table~\ref{tab:opt-drift}; optimizer details and source-subset analyses are in Appendix~\ref{app:optimization-analysis}.

\end{document}